\documentclass{include/latex-class-tlp/tlp}

\usepackage[utf8]{inputenc}
\usepackage{amsmath}
\usepackage{amssymb}
\usepackage{microtype}
\usepackage{listings}
\usepackage{url}

\providecommand{\Underscore}{\textunderscore}

\lstdefinelanguage{clingo}{basicstyle=\ttfamily,keywordstyle=[1]\bfseries,keywordstyle=[2]\bfseries,keywordstyle=[3]\bfseries,showstringspaces=false,literate={_}{\Underscore}1 {\%\%}{}0,escapeinside={\#(}{\#)},alsoletter={\#,\&},keywords=[1]{not,from,import,def,if,else,elif,return,while,break,and,or,for,in,del,and,class,with,as,is,yield,async},keywords=[2]{\#const,\#show,\#minimize,\#base,\#theory,\#count,\#external,\#program,\#script,\#end,\#heuristic,\#edge,\#project,\#show,\#sum},morecomment=[l]{\#\ },morecomment=[l]{\%\ },morestring=[b]",stringstyle={\itshape},commentstyle={\color{darkgray}}}

\lstdefinelanguage{clingcon}[]{clingo}{morekeywords={&dom,&sum,&nsum,&diff,&disjoint,&distinct,&minimize,&maximize,&show}}
\lstdefinelanguage{flingo}[]{clingo}{morekeywords={&sum,&sus,&in,&df,&min,&max,&show}}
\lstdefinelanguage{clingodl}[]{clingo}{morekeywords={&diff}}

\lstdefinelanguage{python}{basicstyle=\ttfamily,keywordstyle=[1]\bfseries,showstringspaces=false,literate={_}{\Underscore}{1},escapeinside={\#(}{\#)},alsoletter={\#,\&},keywords=[1]{not,from,import,def,if,else,elif,return,while,break,and,or,for,in,del,and,class,with,as,is,yield,async},morecomment=[l]{\#\ },morestring=[b]",stringstyle={\itshape},commentstyle={\color{darkgray}}}

 \providecommand{\sysfont}{\textit}

\newcommand{\Clingo}{\sysfont{Clingo}}
\newcommand{\clorm}{\sysfont{clorm}}

\newcommand{\clinguin}{\sysfont{clinguin}}

\newcommand{\aspen}{\sysfont{aspen-tree}}

\newcommand{\clingcon}{\sysfont{clingcon}}
\newcommand{\clingo}{\sysfont{clingo}}

\newcommand{\metasp}{\sysfont{metasp}}

\newcommand{\Telingo}{\sysfont{Telingo}}
\newcommand{\telingo}{\sysfont{telingo}}

\newcommand{\treesitter}{\sysfont{tree-sitter}}

\newcommand{\python}{Python}

 \providecommand{\logfont}{\textrm}

\newcommand{\EL}{\ensuremath{\logfont{EL}}}

\newcommand{\TEL}{\ensuremath{\logfont{TEL}}}

\newcommand{\LDL}{\ensuremath{\logfont{LDL}}}
\newcommand{\LDLf}{\ensuremath{\LDL_{\!f}}}

\newcommand{\DEL}{\ensuremath{\logfont{DEL}}}

\newcommand{\MEL}{\ensuremath{\logfont{MEL}}}

 \RequirePackage{bm}
\RequirePackage{textcomp}
\RequirePackage{upgreek}

\providecommand{\next}{}
\renewcommand{\next}{\text{\rm \raisebox{-.5pt}{\Large\textopenbullet}}}    

\newcommand{\alwaysF}{\ensuremath{\square}}

\newcommand{\eventuallyF}{\ensuremath{\Diamond}}

{}
{}
{}

\newcommand{\finally}{\ensuremath{\bm{\mathsf{F}}}}
\newcommand{\initially}{\ensuremath{\bm{\mathsf{I}}}}

\newcommand{\stp}{\ensuremath{\uptau}}
\mathchardef\mhyphen="2D

\newcommand{\dalways}[1]{\ensuremath{[#1]\,}}                        
\newcommand{\deventually}[1]{\ensuremath{\langle#1\rangle\,}}        \newcommand{\DDia}[1]{\deventually{#1}}

\newcommand{\intervco}[2]{\ensuremath{[#1..#2)}}

\newcommand{\telingolmbd}{\sysfont{telingo-$\lambda$}}

\lstdefinelanguage{clingos}{language=clingo,basicstyle=\small\ttfamily }
\lstdefinelanguage{clingoxs}{language=clingo,basicstyle=\footnotesize\ttfamily }
\makeatletter\lst@AddToHook{OnEmptyLine}{\vspace{\dimexpr-\baselineskip+\smallskipamount\relax}}\makeatother
\makeatletter\def\lst@visiblespace{\lst@ttfamily{~} }\makeatother

\newcommand{\metricI}[2]{\ensuremath{#1_{#2}}}

\newcommand{\light}{\ensuremath{\mathit{l_1}}}

\lefttitle{Susana Hahn et al.}

\jnlPage{1}{8}
\jnlDoiYr{2021}
\doival{10.1017/xxxxx}

\makeatletter
\begin{document}

\title[Temporal Meta-Programming in ASP]{Meta-Programming for Linear-time Temporal Answer Set Programming\thanks{Research partially funded by DFG grant SCHA 550/15, Germany.}}

\begin{authgrp}
  \author{\sn{Susana} \gn{Hahn}}
  \affiliation{University of Potsdam, Germany}\affiliation{Potassco Solutions, Germany}
  \author{\sn{Amad\'e} \gn{Nemes}}
  \affiliation{University of Potsdam, Germany}
  \author{\sn{Javier} \gn{Romero}}
  \affiliation{University of Potsdam, Germany}
  \author{\sn{Torsten} \gn{Schaub}}
  \affiliation{University of Potsdam, Germany}\affiliation{Potassco Solutions, Germany}
\end{authgrp}

\maketitle

\begin{abstract}
The development of temporal extensions of Answer Set Programming (ASP) has led to the emergence of
non-monotonic linear-time (\TEL), dynamic (\DEL), and metric (\MEL) temporal equilibrium logics.
However, the inherent rigidity of highly optimized ASP systems often hinders the rapid exploration and implementation of
alternative logical designs.
In this work, we propose a flexible meta-programming framework that operationalizes the semantics of varied temporal
logics through a unified, declarative framework.
Our approach extends standard ASP meta-programming by augmenting \clingo's theory grammar with formal type
specifications and nesting capabilities.
To ensure semantic correctness, we introduce a transformation pipeline that protects nested modalities from
stable-model-based simplifications during grounding.
We demonstrate the extensibility of our framework by implementing meta-encodings for \TEL, \MEL, and \DEL.
We provide a comprehensive account of \TEL\ and highlight the key features for
managing the interval constraints of \MEL\ and the Fischer-Ladner closure in \DEL.
Finally, we introduce the \metasp\ system,
a versatile tool that encapsulates this workflow.
\end{abstract}
 \section{Introduction}\label{sec:introduction}

The logical foundations of Answer Set Programming (ASP; \citealp{lifschitz19a}) rest upon the Logic of Here-and-There
and its nonmonotonic extension, Equilibrium Logic (\EL; \citealp{pearce06a}).
Hence,
extending ASP often necessitates amalgamating these foundations with complementary formalisms.
In the temporal setting,
this synergy has led to the emergence of
linear-time (\TEL;~\citealp{agcadipescscvi20a}),
dynamic (\DEL;~\citealp{cadilasc20a}), and
metric (\MEL;~\citealp{becadiscsc24a})
temporal equilibrium logics.
While the logic programming fragments of \TEL\ and \DEL\ have been realized in the monolithic \telingo\ system~\citep{cakamosc19a},
meta-encodings for metric variants have only recently emerged~\citep{becadiharosc24a}.
This shift,
moving from hard-coded systems like \telingo\ toward transparent meta-encodings,
is motivated by the architectural rigidity inherent in highly optimized ASP systems.
While such systems excel in performance,
their internal complexity often hinders the rapid implementation of logical extensions.
In contrast,
meta-programming provides an accessible and flexible environment that overcomes this limitation,
enabling the declarative exploration of alternative designs through executable blueprints.

To realize this flexibility,
meta-programming leverages reification:
the process of representing a logic program's syntax as a set of facts.
These facts are combined with a meta-encoding that
implements the desired semantics by
reinterpreting the original program's components.
A major milestone in this area was the introduction of automated reification and a standardized format
in \clingo~\citep{gekasc11b},
initially directed at augmenting optimization
but later applied to broader semantic reinterpretations~\citep{karoscwa21a}.
Notably,
these earlier approaches dealt with the standard ASP language,
ensuring high compatibility with the existing tool-chain.
However, handling foreign language constructs via theory expressions currently introduces a significant ``semantic gap''.
While \clingo\ natively supports their reification,
it produces fine-grained facts representing serialized syntax trees.
Lifting these low-level structures into a format suitable for high-level meta-programming requires substantial manual effort.
Furthermore,
this approach lacks native support for nested theory expressions, restricts their syntactic occurrence,
and has limited integration of syntactic checks on the well-formedness of such expressions.

To address these limitations,
we significantly extend \clingo's meta-programming framework
to support advanced constructs such as temporal modalities, path expressions, and intervals.
By augmenting \clingo's theory grammar with recursive type specifications,
we enable the unrestricted occurrence of these constructs,
including within conditional literals and aggregates.
A fundamental technical challenge in this endeavor is
preventing the grounder from applying standard stable-model simplifications (cf.~Section~\ref{sec:meta}).
Because simplifying nested modalities prematurely compromises semantic correctness,
we introduce a transformation pipeline (Section~\ref{sec:approach}) that insulates these expressions,
preserving the structural integrity of the temporal program.
We detail this approach in the context of \TEL,
and subsequently address the challenges emerging in \MEL\ and \DEL\ in Sections~\ref{sec:mel} and~\ref{sec:del}.
The resulting \metasp\ system (\citealt{metasp}; Section~\ref{sec:system})
is deeply integrated into a modern ASP ecosystem.\footnote{\url{https://potassco.org/{aspen,clinguin,clorm,tree-sitter}}}
By leveraging tools such as \aspen, \clinguin, \clorm, and a custom \treesitter\ grammar,
it provides a robust meta-programming environment for both theoretical exploration and practical application.
Finally,
we evaluate its performance against the monolithic \telingo\ system in Section~\ref{sec:experiments} and
conclude in Section~\ref{sec:discussion}.
 \section{Temporal Logics at a Glance}
\label{sec:temp:glance}

Linear-time temporal logics define temporal models as sequences of states.
Following common conventions in AI~\citep{giavar13a},
we restrict our focus to finite sequences of states,
which serve as the core semantic foundation for the temporal ASP extensions considered in this work.
Standard linear-time logics~\citep{pnueli77a} incorporate temporal modal operators, such as
next (\next), always (\alwaysF), and eventually (\eventuallyF).
Specifically, $\next\varphi$ requires that $\varphi$ holds in the immediate successor state,
while $\alwaysF\varphi$ and $\eventuallyF\varphi$
demand that $\varphi$ holds in all or some future states (including the current one), respectively.
Furthermore, we utilize the initial-state operator \initially,
which holds exclusively in the first state of a sequence.
Additional operators can be derived from these primitives;
e.g.\ the final-state operator $\finally\equiv\neg\next\top$
only holds at the terminal state of a finite sequence.
More formally,
a temporal model $(X_i)_{i=0}^{n}$,
consisting of a sequence of $n{+}1$ interpretations,
satisfies a formula formed with our example operators
at time step $t\in\{0,\dots,n\}$, if
\begin{align}
  (X_i)_{i=0}^{n}, t \models \initially     \quad& \text{ iff } t =0,\label{def:satisfaction:initial}\\
  (X_i)_{i=0}^{n}, t \models \next \varphi       & \text{ iff } t<n \text{ and } (X_i)_{i=0}^{n}, t{+}1 \models \varphi,\label{def:satisfaction:next}\\
  (X_i)_{i=0}^{n}, t \models \eventuallyF\varphi & \text{ iff } (X_i)_{i=0}^{n}, j \models \varphi \text{ for some } j\in\{t,\dots,n\}.\label{def:satisfaction:eventually}
\end{align}

In dynamic logic~\citep{pratt76a},
the modal operators \alwaysF\ and \eventuallyF\ are parametrized by regular expressions
that define paths over sequences of states.
Specifically,
the dynamic formulas $\dalways{\rho}\varphi$ and $\deventually{\rho}\varphi$
require that $\varphi$ holds in all or some of the states, respectively,
reachable via a path segment satisfying the expression $\rho$.
Similarly, in metric temporal logic~\citep{aluhen92a},
\alwaysF\ and \eventuallyF\ are parametrized by a time interval,
which, in our context, is defined over natural numbers.
The metric formulas $\metricI{\alwaysF}{\intervco{l}{u}}\varphi$ and $\metricI{\eventuallyF}{\intervco{l}{u}}\varphi$
stipulate that $\varphi$ holds in all or some states, respectively,
whose associated time point lies within the interval {\intervco{t{+}l}{t{+}u}},
where $t$ denotes the current time point.

While these semantic intuitions have been extensively explored across various logical contexts,
our work centers on adapting them to the non-monotonic realm of Equilibrium Logics.
In particular,
we target fragments of \TEL, \MEL, and \DEL\ that are amenable to implementation via ASP.
To bridge this theoretical foundation with a practical implementation,
we now establish the semantic infrastructure required to represent discrete state sequences
within \clingo's meta-programming framework.

\section{Setting the stage for Temporal Meta-Programming}
\label{sec:meta}

To capture the series of states in linear temporal models,
we present in Listing~\ref{prg:meta:timed:core} a meta-encoding
that extends
the one for plain ASP outlined in~\citep{karoscwa21a}.
\begin{lstlisting}[float=ht,caption={Timed meta-encoding (\texttt{teta.lp})},label={prg:meta:timed:core},language=clingoxs]
time(0..n).%%#(\label{lp:meta:timed:one}#)

conjunction(B,T) :- literal_tuple(B), time(T),%%#(\label{lp:meta:timed:conjunction:one}#)
        hold(L,T) : literal_tuple(B, L), L > 0;
    not hold(L,T) : literal_tuple(B,-L), L > 0.%%#(\label{lp:meta:timed:conjunction:tri}#)

body(normal(B),T) :- rule(_,normal(B)), conjunction(B,T), time(T).%%#(\label{lp:meta:timed:body:normal}#)
body(sum(B,G),T)  :- rule(_,sum(B,G)), time(T),
    #sum { W,L :     hold(L,T), weighted_literal_tuple(B, L,W), L > 0 ;
           W,L : not hold(L,T), weighted_literal_tuple(B,-L,W), L > 0 } >= G.

  hold(A,T) : atom_tuple(H,A)   :- rule(disjunction(H),B), body(B,T), time(T).%%#(\label{lp:meta:timed:head:disjunction}#)
{ hold(A,T) : atom_tuple(H,A) } :- rule(     choice(H),B), body(B,T), time(T).%%#(\label{lp:meta:timed:head:choice}#)
\end{lstlisting}
This extension relies on a time index provided by the atom \lstinline{time(T)} and
the inclusion of the variable \lstinline{T} as the final argument in the encoding's primary predicates.
The set of instantiated rules sharing a common time index effectively duplicates the behavior of the plain ASP meta-encoding.
The number of duplicated ASP encodings is given by the parameter \lstinline{n}; see Line~\ref{lp:meta:timed:one}.

Prior to detailing the encoding in Listing~\ref{prg:meta:timed:core},
we must establish the underlying reified fact format.
For instance, \clingo's reification transforms the (non-temporal) rule
\begin{lstlisting}[language=clingos,numbers=none]
  red(l1) :- not green(l1), light(l1).
\end{lstlisting}
into the following set of facts (omitting definitions for \lstinline{green(l1)} and \lstinline{light(l1)}):
\begin{lstlisting}[language=clingos]
rule(disjunction(0),normal(0)).%%#(\label{lp:reification:one}#)
atom_tuple(0).        atom_tuple(0,3).%%#(\label{lp:reification:two}#)
literal_tuple(0).     literal_tuple(0,-2).  literal_tuple(0,1).%%#(\label{lp:reification:tri}#)
output(light(l1),1).  literal_tuple(1).     literal_tuple(1,1).%%#(\label{lp:reification:output:one}#)
output(green(l1),2).  literal_tuple(2).     literal_tuple(2,2).%%#(\label{lp:reification:output:two}#)
output(red(l1),3).    literal_tuple(3).     literal_tuple(3,3).%%#(\label{lp:reification:output:tri}#)
\end{lstlisting}
The syntactic structure of the rule is captured by the reified facts in Lines~\ref{lp:reification:one}--\ref{lp:reification:tri}.
The primary structure is defined by the instance of \lstinline{rule/2} in Line~\ref{lp:reification:one},
which specifies a disjunctive head and a normal body,
both identified by the constant \lstinline{0}.
In this format,
\clingo\ represents heads and bodies as tuples of atoms and literals, respectively,
utilizing the predicates \lstinline{atom_tuple} and \lstinline{literal_tuple}.
Within these tuples,
ground literals are mapped to unique integer identifiers.
For instance,
the rule's head consists of a single atom represented in Line~\ref{lp:reification:two}
as the tuple \lstinline{(3)},
while the body is represented in Line~\ref{lp:reification:tri}
as a conjunction of two literals forming the tuple \lstinline{(-2,1)}.
Finally,
the \lstinline{output/2} facts in Lines~\ref{lp:reification:output:one}--\ref{lp:reification:output:tri}
establish the necessary mapping
between human-readable symbolic names
and these internal integer identifiers.
For example, \lstinline{output(light(l1),1)} links the symbolic atom to the integer \lstinline{1}
via the corresponding literal tuple \lstinline{(1)}.
\footnote{This indirection is caused by \clingo's normalized show statements that use conditional terms.}

Notably, \clingo's reification process is subject to stable-models-preserving simplifications that occur during grounding.
Consequently, the resulting set of reified facts reflects the simplified rule structure rather than the original program.
If the truth value of a body literal is established during the grounding phase,
that literal is removed from the rule and, by extension, from its reified representation.
For instance, this occurs if
a negative literal `\lstinline{not green(l1)}' is satisfied because \lstinline{green(l1)} is non-derivable.
Conversely,
if a rule's body is determined to be unsatisfiable,
the entire rule becomes vacuous and is eliminated,
along with its associated reified facts.
This scenario arises, for example, if
`\lstinline{not green(l1)}' is found to be false because \lstinline{green(l1)} has been established as true.

Following the reification of the original program,
the resulting facts are combined with a meta-encoding and processed by an ASP system.
In the meta-encoding presented in Listing~\ref{prg:meta:timed:core},
the truth of a literal at a specific time step is represented by instances of the \lstinline{hold/2} predicate.
Accordingly,
the rule in Lines~\ref{lp:meta:timed:conjunction:one}--\ref{lp:meta:timed:conjunction:tri}
derives an instance of \lstinline{conjunction(B,T)} for a time step \lstinline{T},
if all literals within the associated \lstinline{literal_tuple(B)} hold at that time.
This enables the derivation of \lstinline{body(normal(B),T)} in Line~\ref{lp:meta:timed:body:normal}
whenever the literal tuple \lstinline{B} represents a normal rule body.
Consequently, the head atoms of the corresponding rule are derived in
Lines~\ref{lp:meta:timed:head:disjunction} or~\ref{lp:meta:timed:head:choice}.
For instance, using our previous example facts and
assuming that we have \lstinline{time(0)} and \lstinline{hold(1,0)} but not \lstinline{hold(2,0)},
we derive \lstinline{conjunction(0,0)} and \lstinline{body(normal(0),0)},
ultimately concluding \lstinline{hold(3,0)} via the rule in Line~\ref{lp:meta:timed:head:disjunction}.
For a comprehensive account of \clingo's reification, we refer to~\citep{karoscwa21a}.

Thus far,
our meta-encoding captures the program's behavior at each time step independently,
effectively yielding a sequence of isolated state valuations.
In the following,
we describe how temporal operators are utilized to establish the necessary logical connections between these states,
thereby defining the resulting temporal stable models.
 \section{Temporal Meta-Programming}\label{sec:approach}
To illustrate our approach,
we consider a traffic light system modeled with linear-time temporal operators
in \eqref{ex:tel:traffic:light:default} to \eqref{ex:tel:traffic:light:initial}.
In this scenario,
a traffic light is red by default, unless it is green. When a pedestrian button is pushed, the light becomes eventually green from the next step on. The button for light \light\ is pushed one step after the initial state. \begin{align}
  \alwaysF ( \neg\mathit{\mathit{green}(L)} \to \mathit{\mathit{red}(L)} ) \label{ex:tel:traffic:light:default}\\
  \alwaysF ( \mathit{push(L)} \to \next{\eventuallyF\mathit{green(L)}} )   \label{ex:tel:traffic:light:temporal}\\
  \next{\mathit{push(\light)}}                                             \label{ex:tel:traffic:light:initial}
\end{align}
In \TEL,
these constraints yield no temporal models of length one and two,
but one of length three, given by the state sequence:
\(
\langle
\{\mathit{red}(\light)\},
\{\mathit{red}(\light),\mathit{push(\light)}\},
\{\mathit{green(\light)}\}
\rangle
\).
 \subsection{Syntax definition}
To define the additional syntax required for temporal expressions,
we extend the theory language of \clingo\
by \lstinline{#type} definitions.\footnote{\Clingo\ directives are preceded with \lstinline{#}.}
A representative grammar for the type \lstinline{tel} is given in Listing~\ref{prg:tel:grammar},
which serves to augment standard \clingo\ atoms with temporal formulas whose arguments remain regular \clingo\ terms.
\begin{lstlisting}[float=ht,caption={Abridged linear-time temporal grammar (\texttt{tel-grammar.lp})},label={prg:tel:grammar},language=clingos]
#type tel {
    subtypes:    { atom; };                                %%#(\label{tel:grammar:subtype}#)
    expressions: {                                         %%#(\label{tel:grammar:expressions}#)
        &true/0; &not/1: { type: tel, safety: unsafe; };   %%#(\label{tel:grammar:true}#)
        &initial/0;                                        %%#(\label{tel:grammar:initial}#)
        &next/1:         { type: tel, safety:   safe; };   %%#(\label{tel:grammar:next}#)
        &eventually/1:   { type: tel, safety:   safe; }; };%%#(\label{tel:grammar:eventually}#)
    macros:      { &final: &not(&next(&true)); };            %%#(\label{tel:grammar:macro}#)
    occurrence: any;                                       %%#(\label{tel:grammar:any}#)
}.
\end{lstlisting}
This extension is facilitated by the \lstinline{subtypes} field in Line~\ref{tel:grammar:subtype},
permitting the inclusion of other types,
such as the predefined type \lstinline{atom} for standard \clingo\ atoms,
into our temporal expressions.
Consequently, a standard atom like \lstinline{green(l1)} is treated as a \lstinline{tel} expression and
functions as a valid argument of type \lstinline{tel} within nested temporal formulas.
Other supported base types include \lstinline{number}, \lstinline{string}, \lstinline{infimum}, and \lstinline{supremum}.

Lines~\ref{tel:grammar:expressions}--\ref{tel:grammar:macro}
define the formation of temporal \lstinline{expressions}
incorporating the $\top$, $\neg$, \initially, \next, \eventuallyF, and \finally\ operators,
each prefixed by \lstinline{&} in accordance with \clingo’s theory grammar conventions.
To support nesting,
the specification is defined recursively,
allowing constructs such as
\lstinline{&true} or \lstinline{&not} (Line~\ref{tel:grammar:expressions})
to be embedded within other expressions.
This recursive design facilitates the customization of existing constructs and
enables the seamless addition of new operators, such as a Boolean \lstinline{&or}.
Unlike standard theory atoms, these temporal formulas can be nested and
utilized within conditional literals or aggregates.
For example,
the conditional `\lstinline{&eventually(green(L)) : light(L)}'
concisely stipulates that ``every light must eventually turn green.''
Expressions are further refined by specifying
the \lstinline{type} and the \lstinline{safety} condition for their arguments,
which are essential for subsequent transformations.
For instance,
the arguments for \lstinline{&next} and \lstinline{&eventually}
are of type \lstinline{tel} and are declared as \lstinline{safe},
whereas the argument for \lstinline{&not} is declared as \lstinline{unsafe}.
Indeed, the latter declaration can be omitted,
as unspecified \lstinline{safety} conditions default to \lstinline{unsafe}.

Defined operators can be specified with \lstinline{macros}.
In Line~\ref{tel:grammar:macro}, for example,
we define the final-state operator \finally\ as $\neg\next\top$,
which motivates the inclusion of \lstinline{&true} and \lstinline{&not} in the base specification.
For more complex operators with arguments, placeholders can be used,
as detailed in Section~\ref{sec:mel} and~\ref{sec:del}.
Furthermore, the \texttt{occurrence} keyword identifies the valid syntactic positions
for a expressions of a given type,
namely \texttt{any}, \texttt{head}, \texttt{body}, or \texttt{directive}.
While \texttt{tel} expressions are permitted in both heads and bodies of rules (\lstinline{any}),
omitting this declaration restricts a type to serving exclusively as an argument for other expressions.
This distinction is essential for defining auxiliary types, like intervals in Section~\ref{sec:mel},
which act as structural components rather than standalone atoms.

Using the grammar in Listing~\ref{prg:tel:grammar},
we can express the rules in~\eqref{ex:tel:traffic:light:default} to~\eqref{ex:tel:traffic:light:initial}
as in Listing~\ref{prg:tel:example}.
\begin{lstlisting}[float=ht,caption={Linear-time temporal example (\texttt{telex.lp})},label={prg:tel:example},language=clingos]
light(l1).                               %%#(\label{tel:example:one}#)
red(L) :- not green(L), light(L).        %%#(\label{tel:example:two}#)
&next(&eventually(green(L))) :- push(L). %%#(\label{tel:example:tri}#)
&next(push(l1)) :- &initial.             %%#(\label{tel:example:for}#)
\end{lstlisting}
For simplicity, each rule is implicitly governed by the `always' operator \alwaysF,
meaning it applies at every time step of the temporal model.
Rules restricted to the initial state,
like \eqref{ex:tel:traffic:light:initial},
are qualified by the \lstinline{&initial} operator (representing \initially)
within their body to anchor them to the first state of the model.

\subsection{Non-ground Transformations}
Logic programs containing type expressions undergo several transformations to
leverage standard grounding technology without compromising semantic correctness.
The initial phase converts programs with temporal type expressions into standard programs parsable by \clingo.
This transformation is designed to be partially reversible
during output processing to maintain readability.

The primary objective of our non-ground transformations is to avoid incorrect simplifications during grounding.
To achieve this, the system instruments the input program with \lstinline{#external} statements.
As illustrated in Listing~\ref{prg:tel:example:transformed},
\begin{lstlisting}[float=ht,caption={Result of (non-ground) transformation of Listing~\ref{prg:tel:example}},label={prg:tel:example:transformed},language=clingos,firstnumber=6]
#external green(L) : push(L).%%#(\label{tel:example:external:one}#)
#external &initial.          %%#(\label{tel:example:external:two}#)
#external push(l1).          %%#(\label{tel:example:external:tri}#)
\end{lstlisting}
a directive such as \mbox{`\lstinline{#external push(l1).}'} in Line~\ref{tel:example:external:tri}
instructs the grounder to exempt that specific atom from simplifications.
Without this protection,
the grounder would erroneously assume that expressions like
\lstinline{&initial} or instances of \lstinline{push/1} and \lstinline{green/1} are non-derivable and therefore false.
This would lead to semantically incorrect models,
such as a traffic light remaining red indefinitely.

The generation of these protective directives relies on an extended definition of safety.
An atom occurrence in a rule body is deemed \emph{safe}, if it is neither
within the scope of negation nor
within the scope of an argument of a theory expression declared as \lstinline{unsafe} in the grammar.
A rule is considered safe if all its variables occur in safe body atoms.
For example, the rule `\lstinline{wait(L) :- &eventually(green(L)), not green(L).}' is safe
because the variable \lstinline{L} is bound by the occurrence of \lstinline{green(L)} within the first conjunct.
This occurrence is safe because the single argument of \lstinline{&eventually/1}
is declared as \lstinline{safe} in the grammar and is not negated.
Conversely,
if a variable only appears within negated literals,
such as `\lstinline{not green(L)}',
or within theory expressions declared as \lstinline{unsafe} (e.g., \lstinline{&not(green(L))}),
the rule is deemed unsafe.

In our system,
safety verification is interleaved with the generation of external directives.
For each safe rule, \metasp\ generates two distinct types of \lstinline{#external}.
The first type prevents the grounder from discarding rules
with body expressions that might be deemed non-derivable.
The system adds one \lstinline{#external} statement for each temporal expression in the body,
conditioned by the rule's safe body atoms to ensure correct variable binding.
For instance, our \lstinline{wait} example yields `\lstinline{#external &eventually(green(L)): green(L).}'
Similarly,
the statement in Line~\ref{tel:example:external:two} is induced by
the rule in Line \ref{tel:example:for}. The second type ensures that atoms within head expressions are explicitly incorporated into the grounder's instantiation domain.
The system generates an \lstinline{#external} directive
for every atom appearing within a non-negated head expression,
again conditioned on safe body atoms to maintain correct variable bindings.
As illustrated in Line~\ref{tel:example:external:one} of Listing~\ref{prg:tel:example:transformed},
this ensures that atoms like \lstinline{green(L)} remain part of the grounding domain
even when nested within modalities like \lstinline{&eventually}.
This incorporation of \lstinline{#external} directives
allows for a complete ground instantiation in a single call to \clingo,
representing a significant advancement over previous two-pass temporal grounding methods~\citep{capeva09b};
its formal correctness is established in~\citep{nemes24a}.

Declaring \lstinline{safety} conditions for expression arguments is critical to theory grammar design.
A sufficient criterion is whether the satisfaction of a formula implies
the satisfaction of its argument at some time step.
In \TEL,
the arguments of \lstinline{&next/1} and \lstinline{&eventually/1} meet this criterion
and are thus declared \lstinline{safe}.
When this fails, the argument is better declared \lstinline{unsafe}.
For instance,
\cite{nemes24a} shows that the single arguments of $\circ$, $\Diamond$, and $\Box$ provide safe bindings in \TEL\
(as do the second arguments of `until' and `release').
\footnote{Interestingly, this is different in \DEL, where path expressions may interfere
  (cf.\ Listing~\ref{prg:del:grammar} of Section~\ref{sec:del}).}

Finally,
\lstinline{#show} directives are transformed into internal rules and subsequently reified.
This process provides a comprehensive symbol table,
expressed through \lstinline{output/2} facts,
and yields specialized facts such as \lstinline{show_atom/2} and \lstinline{show_term/2} that
capture the syntactic conditions of the original directives.
By leveraging these reified statements,
users can define declarative output policies to customize the output of complex temporal formulas.
We refer the reader to~\citep{metasp} for a detailed account of this mechanism.

\subsection{Extended Reification}
Once non-ground transformations are complete,
the program is grounded and reified using an extension of \clingo's reification process designed for temporal formulas.
During this phase,
type checking and operator expansion are interleaved with grounding,
allowing the system to record each subformula and its associated type
through corresponding \lstinline{formula/2} facts.
In our example,
the resulting subformulas are detailed in
Lines~\ref{tel:example:re:light}--\ref{tel:example:re:initially} of Listing~\ref{prg:tel:example:reified};
\begin{lstlisting}[float=ht,caption={Reified temporal example},label={prg:tel:example:reified},language=clingoxs,firstline=55, lastline=67]
output(light(l1),6).    output(&next(push(l1)),7).                %%#(\label{tel:example:symbols:one}#)
output(green(l1),2).    output(&next(&eventually(green(l1))),8).
output(red(l1),4).      output(&initial,5).
output(push(l1),1).                                                %%#(\label{tel:example:symbols:for}#)

formula(atom,light(l1)).             formula(tel,light(l1)). %%#(\label{tel:example:re:light}#)
formula(atom,red(l1)).               formula(tel,red(l1)).
formula(atom,push(l1)).              formula(tel,push(l1)).
formula(atom,green(l1)).             formula(tel,green(l1)). %%#(\label{tel:example:re:green}#)
formula(tel,&eventually(green(l1))).                        %%#(\label{tel:example:re:eventually}#)
formula(tel,&next(&eventually(green(l1)))).                %%#(\label{tel:example:re:next:eventually}#)
formula(tel,&next(push(l1))).
formula(tel,&initial).                                      %%#(\label{tel:example:re:initially}#)
\end{lstlisting}
for instance, Lines~\ref{tel:example:re:green}--\ref{tel:example:re:next:eventually}
represent the reified subexpressions derived from \lstinline{&next(&eventually(green(l1)))}.
Notably, a single subformula may appear in multiple \lstinline{formula/2} facts, depending on its type memberships.
For example,
\lstinline{green(l1)} is simultaneously categorized as an \lstinline{atom} and a temporal expression (\lstinline{tel}).
The \lstinline{formula/2} predicate serves as a domain predicate in our meta-encodings,
effectively constraining the instantiation of formula-based rules.
This representation allows us to easily incorporate advanced syntactic concepts,
such as the closure of~\cite{fislad79a} in Section~\ref{sec:del},
by simply adding auxiliary rules to define extended subexpression relationships.

Standard symbol table information is retained via the \lstinline{output/2} predicate.
Because our non-ground transformations instrument the program
by inserting \lstinline{#external} directives and removing \lstinline{#show} statements,
the grounder is prevented from aggressively simplifying the program and
all atoms and temporal formulas are preserved as \lstinline{output/2} facts.
The symbol table obtained in our example is given in
Lines~\ref{tel:example:symbols:one}--\ref{tel:example:symbols:for}
of Listing~\ref{prg:tel:example:reified}.

The core of this reified representation utilizes standard reification predicates:
a \lstinline{rule/2} instance for each rule in the original program is provided alongside
the associated \lstinline{atom_tuple} and \lstinline{literal_tuple} facts that
capture their respective heads and bodies
as outlined in Section~\ref{sec:meta}.
In addition to the \lstinline{output/2} and \lstinline{formula/2} facts,
the representation incorporates
\lstinline{external/2} facts, as well as
\lstinline{show_term/2} and \lstinline{show_atom/2} instances.
These correspond to the injected \lstinline{#external} directives and transformed \lstinline{#show} statements.
For the sake of brevity, these specific predicate instances are omitted from the discussion.

\subsection{Temporal Meta-encoding}
A meta-encoding operationalizes the semantics of a logic program
by defining the underlying semantic structure and interpreting the program's components within that structure.
The design of our meta-encoding reflects this bipartite nature:
the first part (Listing~\ref{prg:meta:timed:core})
establishes a sequence of independent stable models,
where temporal formulas are treated as standard atoms,
while the second (Listing~\ref{prg:tel:semantics})
accounts for the satisfaction relation of temporal operators \initially, \next, and \eventuallyF.

The first part of our meta-encoding employs the predicate \lstinline{hold/2}
to represent the valuation of literals along different time steps;
this representation captures linear-time temporal models and is thus common to the temporal logics \TEL, \MEL, and \DEL.

The transition between both parts relies on the \lstinline{true/2} predicate.
While \lstinline{hold/2} refers to atoms and formulas via numeric identifiers,
\lstinline{true/2} utilizes their symbolic representations.
This distinction is vital,
as it enables the evaluation of operators based on their inherent syntactic structure rather than opaque identifiers.
To this end,
the \lstinline{hold/2} atoms
are related to nested temporal formulas through the \lstinline{true/2} predicate
in Listing~\ref{prg:tel:semantics}
to capture the inductive definition of satisfaction.
\begin{lstlisting}[float=th,caption={Encoding of \TEL\ semantics of\/ \initially, \next, and \eventuallyF\ (\texttt{tel-semantics.lp})},label={prg:tel:semantics},language=clingoxs,breaklines=false,showspaces=true]
true(O,T) :- output(O,B), time(T), hold(L,T): literal_tuple(B,L).%%#(\label{lp:meta:timed:true:hold}#)
hold(L,T) :- output(O,B), time(T), true(O,T), literal_tuple(B,L).%%#(\label{lp:meta:timed:hold:true}#)

true(&initial,T) :- formula(tel,&initial), T = 0.                               %%#(\label{enc:satisfaction:initial:one}#)
           T = 0 :- formula(tel,&initial), true(&initial,T).                                         %%#(\label{enc:satisfaction:initial:two}#)

true(&next(F),T) :- formula(tel,&next(F)), T < n, true(F,T+1).                          %%#(\label{enc:satisfaction:next:one}#)
           T < n :- formula(tel,&next(F)), true(&next(F),T).                       %%#(\label{enc:satisfaction:next:two}#)
      true(F,T+1):- formula(tel,&next(F)), true(&next(F),T).                       %%#(\label{enc:satisfaction:next:tri}#)

  true(F,J) : J=T..n :- formula(tel,&eventually(F)), true(&eventually(F),T).         %%#(\label{enc:satisfaction:eventually:one}#)
true(&eventually(F),T) :- formula(tel,&eventually(F)),
                          J=T..n, true(F,J), time(T). %%#(\label{enc:satisfaction:eventually:two}#)
\end{lstlisting}
The rules in Lines~\ref{lp:meta:timed:true:hold}--\ref{lp:meta:timed:hold:true}
establish the base case by
synchronizing \lstinline{hold/2} and \lstinline{true/2} instances for all elements in the symbol table.
Whenever a symbolic representation \lstinline{o} corresponds to a numeric identifier \lstinline{l},
the system typically relies on the presence of the facts \lstinline{output(o,b)} and \lstinline{literal_tuple(b,l)}
to enforce the equivalence between \lstinline{true(o,t)} and \lstinline{hold(l,t)}
along all time steps \lstinline{t}.
However, a subtle distinction arises when a formula \lstinline{o} appears as a fact:
because it lacks an associated rule body,
the corresponding \lstinline{literal_tuple/2} is absent from the reified output.
This is elegantly handled by the rule in Line~\ref{lp:meta:timed:true:hold}
via a conditional,
which allows for deriving \lstinline{true(o,t)} directly from the symbol table,
while the first part of the meta-encoding independently establishes the truth of the corresponding fact \lstinline{hold(l,t)}.

Building on this,
the satisfaction relations for \initially, \next, and \eventuallyF\ are
implemented by the subsequent rules in Listing~\ref{prg:tel:semantics}.
The individual cases, and thus the overall grounding process, is governed by \lstinline{formula/2},
acting as a domain predicate.
Each group of rules in Lines~\ref{enc:satisfaction:initial:one}--\ref{enc:satisfaction:initial:two},
\ref{enc:satisfaction:next:one}--\ref{enc:satisfaction:next:tri}, and~\ref{enc:satisfaction:eventually:one}--\ref{enc:satisfaction:eventually:two}
establishes a logical equivalence that captures the formal truth conditions for \initially, \next, and \eventuallyF\
as defined in
\eqref{def:satisfaction:initial},
\eqref{def:satisfaction:next}, and
\eqref{def:satisfaction:eventually},
respectively.
For instance, the satisfaction of $\eventuallyF\varphi$ in \eqref{def:satisfaction:eventually}
can be expressed in our framework as
\lstinline{true(&eventually(F),T)} iff \lstinline{true(F,j)} for some $\mathtt{j} \in \{\mathtt{T},\dots,\mathtt{n}\}$.
This bidirectional relationship is realized in
Lines~\ref{enc:satisfaction:eventually:one} and~\ref{enc:satisfaction:eventually:two},
whose rules collectively enforce the semantics of the `eventually' operator.

Taken together,
the theory grammar in Listing~\ref{prg:tel:grammar} and
the meta-encodings in Listing~\ref{prg:meta:timed:core} and~\ref{prg:tel:semantics} provide a complete framework for
evaluating our temporal program presented in Listing~\ref{prg:tel:example}.
More generally, given a temporal logic program $P$, and $n\geq 0$,
we get a 1--1 correspondence between
the temporal models of length $n$ of $P$ and
the stable models of the reification of $P$, obtained using Listing~\ref{prg:tel:grammar},
together with Listings~\ref{prg:meta:timed:core} and~\ref{prg:tel:semantics}.
The broader functionality of our tool, \metasp,
is discussed in Section~\ref{sec:system}.

 \section{Metric Meta-Programming}
\label{sec:mel}

Having established our meta-programming workflow through \TEL,
we now demonstrate the framework's extensibility within the more complex setting of \MEL.
This section illustrates how our approach handles quantitative temporal constraints
by leveraging a modular grammar and hybrid solving techniques.

The grammar for this metric extension, detailed in Listing~\ref{prg:mel:grammar},
\begin{lstlisting}[float=ht,caption={Abridged metric temporal grammar},label={prg:mel:grammar},language=clingoxs,breaklines=false,showspaces=true]
#type ub { subtypes: { number; supremum; }; }.                                      %%#(\label{mel:grammar:ub}#)
#type interval { expressions: { &i/2: { type: number; type: ub; }; }; }.            %%#(\label{mel:grammar:interval}#)
#type mel {
 subtypes:     { atom; };                                                           %%#(\label{mel:grammar:subtype}#)
 expressions:  {                                                                    %%#(\label{mel:grammar:expressions}#)
  &initial/0;                                                                       %%#(\label{mel:grammar:initial}#)
  &next/2:       {type: interval, safety: unsafe; type: mel, safety: safe;};      %%#(\label{mel:grammar:next}#)
  &eventually/2: {type: interval, safety: unsafe; type: mel, safety: safe;};      %%#(\label{mel:grammar:eventually}#)
 };
 macros:       { &next(v): &next(&i(0,#sup), v); where: {v: mel;}; };               %%#(\label{mel:grammar:macro}#)
 occurrence: any;                                                                   %%#(\label{mel:grammar:occurrence}#)
}.
\end{lstlisting}
incorporates the auxiliary types \lstinline{interval} and \lstinline{ub}
to facilitate the representation of temporal intervals.
Intervals are constructed using the binary operator \lstinline{&i},
which pairs a numeric lower bound with an upper bound of type \lstinline{ub} (Line~\ref{mel:grammar:interval}).
As specified in Line~\ref{mel:grammar:ub},
this upper bound may be either a specific \lstinline{number} or
the constant \lstinline{#sup},
the sole member of the predefined \lstinline{supremum} type.
Both types are predefined in \metasp\ and serve purely as structural components.
In fact,
while expressions of type \lstinline{mel} are permitted to occur in rule heads and bodies
(as declared in Line~\ref{mel:grammar:occurrence}),
the \lstinline{interval} and \lstinline{ub} types are restricted from appearing as standalone atoms.
This is enforced by the absence of an \lstinline{occurrence} declaration for these types.
Furthermore,
both arguments of the interval operator \lstinline{&i} are considered \lstinline{unsafe},
as reflected by the lack of explicit \lstinline{safety} declarations in the grammar.

Following the pattern of \TEL\ expressions,
Line~\ref{mel:grammar:expressions} to~\ref{mel:grammar:eventually}
define formulas involving expressions with
\initially, $\metricI{\next}{\intervco{l}{u}}\varphi$, and $\metricI{\eventuallyF}{\intervco{l}{u}}\varphi$.
The latter two are captured by the binary operators \lstinline{next/2} and \lstinline{eventually/2},
which pair a metric \lstinline{interval} with a \lstinline{mel} formula.
Similar to \TEL, the second argument of type \lstinline{mel} is considered \lstinline{safe},
while the first, \lstinline{interval} argument is declared \lstinline{unsafe}.
To streamline the syntax,
Line~\ref{mel:grammar:macro} introduces a unary variant of the \lstinline{&next} operator
representing the unconstrained metric operator $\metricI{\next}{\intervco{0}{\omega}}$.
In this context, the interval imposes no constraints, and the operator amounts to its qualitative counterpart \next\ in \TEL.
This allows us to write \lstinline{&next(F)} as a shorthand for the more verbose \lstinline{&next(&i(0,#sup),F)},
where \lstinline{F} is a \lstinline{mel} formula.
To facilitate this in the theory specification, placeholders are utilized;
these are specific constants that are declared and assigned types within a \lstinline{where} clause.

The metric counterpart of our traffic lights example is provided in Listing~\ref{prg:mel:example}.
\begin{lstlisting}[float=ht,caption={Metric temporal example},label={prg:mel:example},language=clingos]
light(l1).
red(L) :- not green(L), light(L).
&eventually(&i(10,15),green(L)):- push(L). %%#(\label{melex:eventually}#)
&next(push(l1)) :- &initial. %%#(\label{melex:next}#)
\end{lstlisting}
This encoding differs from Listing~\ref{prg:tel:example} primarily in Line~\ref{melex:eventually},
where the plain eventually head atom is replaced by its metric counterpart \lstinline{&eventually(&i(10,15),green(L))},
stipulating that the light must turn green between time points 10 and 15.
The occurrence of \lstinline{&next(push(l1))} in Line~\ref{melex:next} is replaced during reification by \lstinline{&next(&i(0,#sup),push(l1))}.

To accommodate \MEL's dual nature,
combining qualitative temporal progression with quantitative timing constraints,
our meta-encoding utilizes the hybrid ASP solver \clingcon~\citep{bakaossc16a}
to leverage its native support for linear constraints over integers.
Formally,
each state $X_t$ in a temporal model $(X_i)_{i=0}^{n}$ is associated in \MEL~\citep{becadiscsc24a}
with a discrete time point via a timing function $\tau$
satisfying
$\tau(0) = 0$ and
$\tau(t) - \tau(t-1) \geq 1$ for $t \in \{1,\ldots,n\}$.
This relationship is operationalized in Listing~\ref{prg:mel:tau} using \clingcon's \lstinline{&sum} theory atom,
in which each instance of functor \lstinline{t/1} takes a non-negative integer value
corresponding to a specific time point.
\begin{lstlisting}[float=ht,caption={Encoding of the timing function $\tau$},label={prg:mel:tau},language=clingos, firstline=1, lastline=2]
&sum{ t(0) } = 0. %%#(\label{lp:mel:sum:zero}#)
&sum{ t(T); -t(T-1) } >= 1 :- T = 1..n. %%#(\label{lp:mel:sum:one}#)
\end{lstlisting}

We illustrate the meta-encoding of metric operators in terms of the metric next $\metricI{\next}{\intervco{l}{u}}$.
As with \TEL,
the encoding is derived from semantic principles:
A temporal model $(X_i)_{i=0}^{n}$ with a timing function $\tau$
satisfies $\metricI{\next}{\intervco{l}{u}} \varphi$ at time step $t\in\{0,\dots,n\}$,
written
\(
(X_i)_{i=0}^{n}, t \models \metricI{\next}{\intervco{l}{u}} \varphi
\),
if
\begin{align*}
  (X_i)_{i=0}^{n},
  t{+}1 \models \varphi \text{ and }
  \tau(t+1)-\tau(t)\in \intervco{l}{u}
  \text{ for }
  t<n
  \ .
\end{align*}
We can express this within our framework as:
\lstinline{true(&next(&i(L,U),F),T)} iff
  \lstinline{true(F,T+1)},
  \lstinline|&sum { t(T+1); -t(T) } >= L|,
  \lstinline|&sum { t(T+1); -t(T) } < U|, and
  \lstinline{T < n}.
The right-to-left direction of this equivalence is given in Listing~\ref{prg:mel:next}.
\begin{lstlisting}[float=ht,caption={Encoding for MEL semantics of \metricI{\next}{\intervco{l}{u}}},label={prg:mel:next},language=clingoxs, firstline=4, lastline=6,breaklines=false,showspaces=true]
true(&next(&i(L,U),F),T) :- formula(mel,&next(&i(L,U),F)), time(T), %%#(\label{enc:satisfaction:mel:next:l2r:start}#)
  true(F,T+1), &sum{t(T+1); -t(T)} >= L, &sum{t(T+1); -t(T)} < U, T < n. %%#(\label{enc:satisfaction:mel:next:l2r:end}#)
\end{lstlisting}
The left-to-right direction is encoded analogously with four rules, one for each conjunct.
The full encoding can be found at~\citep{metasp}.
 \section{Dynamic Meta-Programming}
\label{sec:del}

Lastly, we demonstrate the extensibility of our approach through \DEL.
The grammar for this extension, provided in Listing~\ref{prg:del:grammar},
inherits the \TEL\ type definitions via the \lstinline{subtypes} field.\footnote{Alternatively, \lstinline{del} and \lstinline{tel} types can be defined next to each other
  (or even one redefined in the other).}
Path expressions, defined in Lines~\ref{del:grammar:expressions:path}--\ref{del:grammar:macro},
serve exclusively as arguments for the dynamic modal operators.
These paths incorporate standard constructors,
such as step ($\stp$), test ($?$), star ($*$), choice ($+$), and sequence ($;$),
to define regular expressions over state sequences.
\begin{lstlisting}[float=ht,caption={Abridged dynamic temporal grammar},label={prg:del:grammar},language=clingoxs,breaklines=false,showspaces=true]
#type del{
 subtypes:    { tel; };                                                      %%#(\label{del:grammar:subtype}#)
 expressions: {                                                              %%#(\label{del:grammar:expressions:del}#)
  &always/2:     {type: path, safety: unsafe; type: del, safety: unsafe;};   %%#(\label{del:grammar:always}#)
  &eventually/2: {type: path, safety:   safe; type: del, safety:   safe;};};%%#(\label{del:grammar:eventually}#)
 occurrence:  any;                                                           %%#(\label{del:grammar:any}#)
}.
#type path {
 expressions: {                                                              %%#(\label{del:grammar:expressions:path}#)
  &step/0;                                                                   %%#(\label{del:grammar:step}#)
  &test/1:   { type: del,  safety:   safe; };                                %%#(\label{del:grammar:test}#)
  &star/1:   { type: path, safety: unsafe; };                                %%#(\label{del:grammar:star}#)
  &choice/2: { type: path, safety: unsafe; type: path, safety: unsafe; };    %%#(\label{del:grammar:choice}#)
  &seq/2:    { type: path, safety:   safe; type: path, safety:   safe; }; }; %%#(\label{del:grammar:seq}#)
 macros:      {  a: &seq(&test(a), &step); where: { a: atom; }; };           %%#(\label{del:grammar:macro}#)
}.
\end{lstlisting}The macro introduced in Line~\ref{del:grammar:macro} permits an atom $a$ to be used directly in path expressions
by translating it into the expression $(a?;\stp)$,
a replacement strictly confined to the context of \lstinline{path} expressions.\footnote{Note that \DEL\ follows the single-sorted approach of \LDLf~\citep{giavar13a}.}

For example, the formula
\(
\initially\wedge\DDia{(\mathit{green(l_1)};\mathit{red(l_1)})^*} \finally
\)
describes temporal models where light $l_1$ alternates between green and red.
This is expressed in our framework as an integrity constraint:
\begin{lstlisting}[numbers=none,language=clingos]
:- &initial, not &eventually(&star(&seq(green(L),red(L))),&final).
\end{lstlisting}
This constraint effectively utilizes the \lstinline{tel} operators \lstinline{&initial} and \lstinline{&final}
alongside the new dynamic constructs.
The instances of \lstinline{green(L)} and \lstinline{red(L)} are expanded during reification into
\lstinline{&seq(&test(green(L)),&step)} and \lstinline{&seq(&test(red(L)),&step)}.

In \DEL,
the satisfaction relation has a scope that extends beyond the subformulas explicitly present in the input.
One must also capture the recursive interplay between dynamic modal operators and path expressions,
which requires a superset of the subformulas referred to as the \cite{fislad79a} closure.
In our framework,
this closure is obtained declaratively by extending the definition of the \lstinline{formula/2} predicate,
as demonstrated for \lstinline{&eventually} in Listing~\ref{prg:del:closure}.
\begin{lstlisting}[float=ht,caption={Additional rules for the Fischer-Ladner closure of \texttt{\&eventually} formulas},label={prg:del:closure},language=clingo,basicstyle=\scriptsize\ttfamily,firstline=7]
formula(del,&eventually(P,&eventually(&star(P),F))) :- formula(del,&eventually(&star(P),F)).%%#(\label{del:flc:eventually:star}#)
formula(del,&eventually(P1,F)) :- formula(del,&eventually(&choice(P1,P2),F)).
formula(del,&eventually(P2,F)) :- formula(del,&eventually(&choice(P1,P2),F)).
formula(del,&eventually(P1,&eventually(P2,F))) :- formula(del,&eventually(&seq(P1,P2),F)).%%#(\label{del:flc:eventually:seq:one}#)
formula(del,&eventually(P2,F)) :- formula(del,&eventually(&seq(P1,P2),F)).%%#(\label{del:flc:eventually:seq:two}#)
\end{lstlisting}
For instance,
Line~\ref{del:flc:eventually:star} unfolds the `star' operator by one repetition, while
Line~\ref{del:flc:eventually:seq:one} decomposes a sequence into nested modalities.
This declarative expansion of the subformulas ensures that the meta-encoding can verify path satisfaction
through a finite set of sub-expressions derived entirely via ASP rules.
The full encoding can be found at~\citep{metasp}.
 \section{The \metasp\ System}
\label{sec:system}

The \metasp\ system encapsulates our meta-programming workflow,
providing a unified interface for defining and solving temporal ASP extensions.
The system is implemented via an \lstinline{Application} class that extends the target system's command line,
ensuring full access to native options, such as optimization.

The \lstinline{solve} command orchestrates the entire workflow from syntax parsing to model generation.
\begin{lstlisting}[float=ht,language=bash,basicstyle=\small\ttfamily,caption={System output showcasing the \TEL\ setting},label={prg:tel:example:output},lastline=12]
UNIX> metasp solve clingo telex.lp                    \
      --syntax-encoding tel-grammar.lp                \
      --semantics-encoding teta.lp tel-semantics.lp   \
      --printer temporal_printer -c n=2
Reading from telex.lp
Solving...
Answer: 1
 State 0: light(l1) red(l1)
 State 1: light(l1) push(l1) red(l1)
 State 2: green(l1) light(l1)

SATISFIABLE
\end{lstlisting}
As shown in Listing~\ref{prg:tel:example:output},
it requires a target solver,
typically \clingo, or \clingcon\ for metric extensions,
alongside problem-specific and logic-specific encodings.

In Listing~\ref{prg:tel:example:output},
we \lstinline{solve} our temporal example in file \lstinline{telex.lp} (Listing~\ref{prg:tel:example}) with \lstinline{clingo}.
The options \lstinline{--syntax-encoding} and \lstinline{--semantics-encoding}
are used to provide the grammar and meta-encodings for the desired logic.
Users can customize the output via the \lstinline{--printer} option.
For example,
the built-in \lstinline{temporal_printer} organizes atoms by state to improve readability of temporal models.
If omitted, the target system's default printer is used,
though users may also provide custom \python\ scripts for specialized output policies.
Standard \clingo\ arguments, such as `\lstinline{-c n=2}' for assigning constant values, are supported directly.
The output in Listing~\ref{prg:tel:example:output} reflects a single temporal model,
where the light becomes \textit{green} at state 2, as expected.

To facilitate development,
\metasp\ allows users to intercept the transformation process at individual stages: the
\lstinline{transform} command prints the given program after initial first-order transformations, while
\lstinline{reify} outputs the extended reification facts.
Unlike \lstinline{solve},
these diagnostic commands require only grammar definitions and constants, bypassing the need for a back-end solver.
For interactive solution exploration,
the \lstinline{ui} command launches an interface using \lstinline{clinguin}~\citep{behasc24a}.
To maintain a clean command-line experience,
extension-specific options can be aggregated into configuration files.
Finally, \metasp\
supports multiple log levels and issues proactive warnings for common development issues,
such as type mismatches or macro redefinitions.
 \section{Experiments}
\label{sec:experiments}

We conducted an empirical evaluation to assess the performance trade-offs between
the flexible meta-programming framework of \metasp\ and
the specialized architecture of \telingo.
Our comparison focuses on \TEL\ and \DEL, as these are currently the only temporal extensions supported by \telingo.
All experiments were conducted on Debian Linux~10 machines with an Intel Xeon E5-2650v4 CPU (2.9 GHz) and 64 GB of RAM.
We compare \metasp~1.0, utilizing the meta-encodings and grammars partly introduced above with \clingo~5.8,
against a specialized fork of \telingo~2.1.3.
Individual runs were constrained by a time limit of 20
minutes and a memory limit of 24GB.

\Telingo\ solves \TEL\ and \DEL\ problems incrementally, gradually expanding the horizon.
Because our meta-encodings evaluate a fixed horizon,
the repeated unsatisfiability checks of the incremental approach would obscure a direct performance comparison.
Therefore, we developed \telingolmbd,
a specialized fork of \telingo\ that supports fixed-horizon solving.\footnote{Available at \url{https://potassco.org/telingo}.}

\begin{figure}[ht]
  \includegraphics[width=0.45\textwidth]{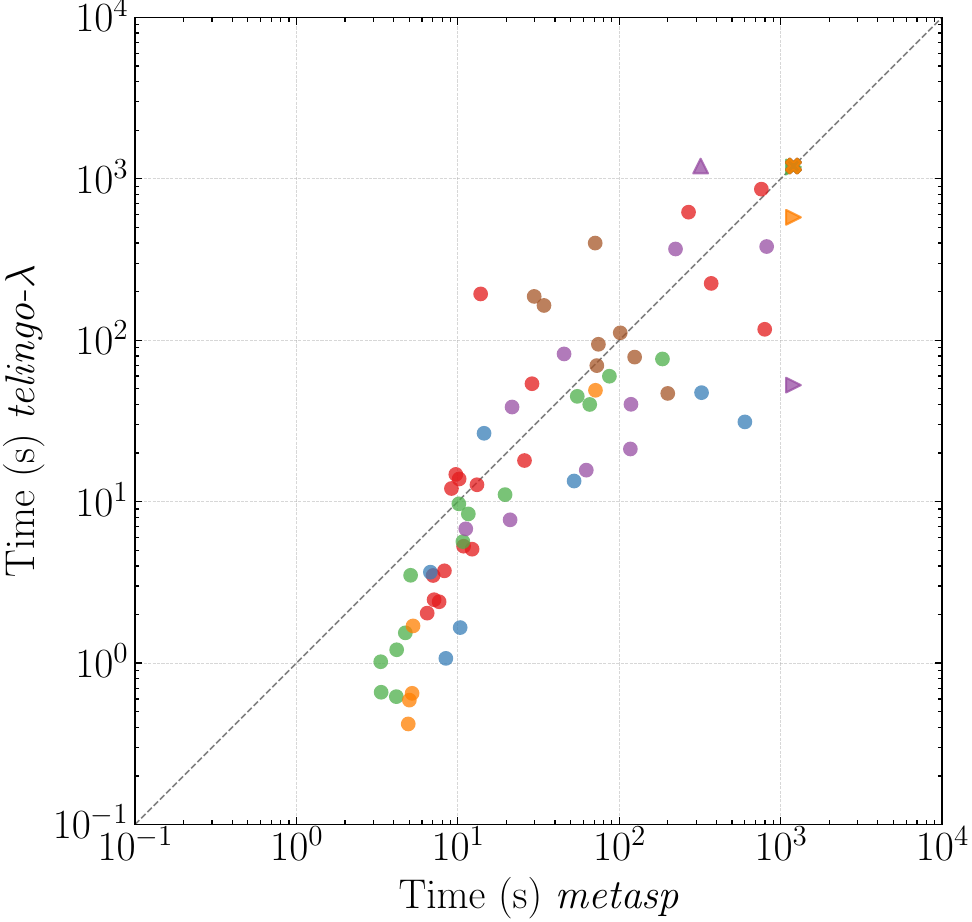}
  \includegraphics[width=0.45\textwidth]{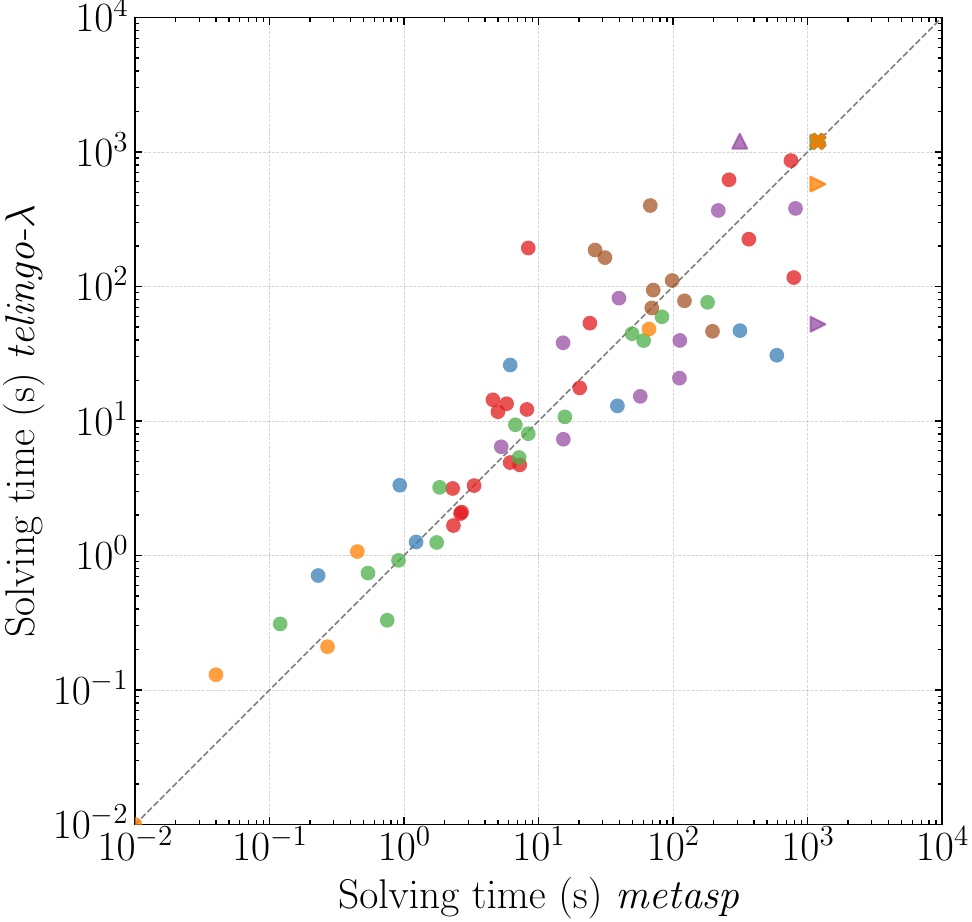} \caption{Comparison of runtime and solving time of \metasp\ vs \telingolmbd\ on \TEL\ problems.}
  \label{fig:telingo-metasp-exper}
\end{figure}
Our \TEL\ evaluation utilizes 69 temporal planning instances from~\citep{cakamosc19a}, spanning classic domains:
\textit{hanoi-towers} (red),
\textit{labyrinth} (blue),
\textit{no-mystery} (green),
\textit{richochet-robots} (purple),
\textit{sokoban} (orange), and
\textit{visitall} (brown).
The results are visualized in Figure~\ref{fig:telingo-metasp-exper},
where timeouts for \telingolmbd\ and \metasp\ are indicated by upward and rightward arrows, respectively;
crosses indicate joint timeouts.

A clear performance trend emerges:
on easier instances, \telingolmbd\ exhibits faster total runtimes,
but as complexity increases, the systems perform competitively.
The solving time graph (right) clarifies this discrepancy. Here, the systems compete head-to-head, indicating the runtime difference is primarily driven by an initial grounding overhead in \metasp.
This penalty is expected, as \metasp\ requires two distinct grounding phases: first to reify the input program, and second to instantiate the meta-encoding.
Crucially, because time is not explicitly represented in \TEL, the initial reification step does not scale with the horizon.
Consequently, this overhead remains relatively flat and becomes negligible on more difficult instances requiring longer horizons.

Next, we evaluate \metasp\ in the context of \DEL\ using the elevator benchmark from~\citep{lereleli97a}.
While we adopt the general experimental setup of~\citet{cadilasc20a},
our evaluation employs a more recent version of \telingo,
which accounts for variations in the reported statistics.
This experiment analyzes the impact of \DEL-based constraints,
designed to eliminate redundant elevator movements,
on the number of temporal traces, search choices, and final solver constraints.
The full experiment comprises several elevator configurations,
serving 5, 7, 9, and 11~floors, each evaluated over five different horizons.
To illustrate the different behaviors of the systems,
we discuss one representative instance in detail:
the 11-floor configuration, which admits two optimal traces of length~17,
evaluated at horizon~21.
In the baseline configuration without control rules, the instance yields 200900 temporal traces.
To compute these,
\telingolmbd\ and \metasp\ explore  search spaces of comparable size (216908 versus 214840 choices) and
produce a similar number of constraints (3221 versus 3508).
Upon introducing the \DEL\ control rule,
the number of traces drops to 2, aligning with the aforementioned optima.
While both systems see a massive search space reduction,
\telingolmbd\ prunes more aggressively (to 50 choices) than \metasp\ (192 choices).
However, this optimization incurs a structural cost for both:
the constraint count increases from 3221 to 5058 for \telingolmbd\ and from 3508 to 6086 for \metasp.
This behavior, consistent across the benchmark suite, highlights a fundamental key trade-off:
both systems achieve a drastic search space reduction at the expense of a significantly inflated grounded representation.

While these findings are promising,
a broader empirical and theoretical investigation is necessary to fully characterize this behavior across diverse problem domains.
Detailed results for all experiments are available in the \metasp\ repository~\citep{metasp}.
 \section{Conclusion}\label{sec:discussion}

We have introduced a meta-programming framework that
facilitates the declarative implementation of diverse linear-time temporal logics.
By enabling the rapid exploration of alternative logical designs without requiring modifications to the underlying solver,
our approach provides a versatile environment suited for both theoretical research and the generation of system blueprints,
even beyond the domain of temporal logics.

The framework's efficacy rests on two primary pillars:
an extension of the \clingo\ theory grammar for temporal expressions and
specialized program transformations that insulate nested formulas
from the grounder's standard simplifications.
From a technical perspective,
we have introduced an extended concept of safety that is more permissive than the one established in~\citep{agcapevidi17a}.
Furthermore, our strategic use of \lstinline{#external} directives enables the grounding of temporal programs in a single pass,
a significant advancement over previous two-pass temporal grounding methods~\citep{agcapevidi17a},
bridging the gap between high-level temporal specifications and standard ASP grounding technology.
Moreover,
we have demonstrated that logic-specific meta-encodings are directly derivable from a logic's satisfaction relation,
ensuring semantic clarity and lowering the barrier for proving completeness and correctness,
as done by~\cite{nemes24a}.
Unlike earlier approaches that focus on individual arguments~\citep{bagezh13a},
our typing mechanism targets atomic expressions themselves.

While our current meta-encodings rely on explicit discrete time steps,
several promising avenues for future development remain.
We aim to explore automata-based computations \citep{cabdem11a,cumaperi24a} and
the integration of acyclicity and difference constraints \citep{bekascsosvwa24a}.
Furthermore,
we seek to determine how direct temporal encodings \citep{helnie03a,fiieri24a}
can be integrated within our declarative meta-programming environment.
Ultimately,
by bridging the gap between high-level logical specifications and low-level solver performance,
the \metasp\ system serves as a flexible and robust platform for the future development of temporal ASP extensions and beyond.
 \bibliographystyle{plainnat}

\end{document}